\documentclass[11pt,a4paper]{article}
\usepackage[hyperref]{acl2020}
\usepackage{times}
\usepackage{latexsym}
\usepackage{graphicx}

\usepackage{algorithm}
\usepackage{algorithmic}
\usepackage{float,mathtools}

\usepackage{multirow}
\usepackage{listings}

\usepackage{microtype}

\aclfinalcopy 
\def\aclpaperid{2885} 

\title{schuBERT: Optimizing Elements of BERT}

\author{Ashish Khetan \\
  Amazon AWS \\
  \texttt{khetan@amazon.com} \\\And
  Zohar Karnin \\
 Amazon AWS \\
  \texttt{zkarnin@amazon.com} \\}

\date{}

\begin{document}
\maketitle
\begin{abstract}
Transformers \citep{vaswani2017attention} have gradually become a key component for many state-of-the-art natural language representation models. A recent Transformer based model- BERT \citep{devlin2018bert} achieved state-of-the-art results on various natural language processing tasks, including GLUE, SQuAD v1.1, and SQuAD v2.0. This model however is computationally prohibitive and has a huge number of parameters. In this work we revisit the architecture choices of BERT in efforts to obtain a lighter model. We focus on reducing the number of parameters yet our methods can be applied towards other objectives such FLOPs or latency. We show that much efficient light BERT models can be obtained by reducing algorithmically chosen correct architecture design dimensions rather than reducing the number of Transformer encoder layers. In particular, our schuBERT gives $6.6\%$ higher average accuracy on GLUE and SQuAD datasets as compared to BERT with three encoder layers while having the same number of parameters. 
\end{abstract}

\section{Introduction}
Transformer \citep{vaswani2017attention} based models have achieved state-of-the-art performance for many natural language processing tasks \citep{dai2015semi, peters2018deep, radford2018improving, howard2018universal}. These include machine translation \citep{vaswani2017attention, ott2018scaling}, question-answering tasks \citep{devlin2018bert}, natural language inference \citep{bowman2015large, williams2017broad} and semantic role labeling \citep{strubell2018linguistically}. 

A recent Transformer based model BERT \citep{devlin2018bert} achieved state-of-the-art results on various natural language processing tasks including GLUE, SQuAD v1.1 and SQuAD v2.0. BERT's model architecture is a multi-layer bidirectional Transformer 
encoder based on the original implementation described in \citet{vaswani2017attention}.

Following the seminal results obtained by the BERT model, several follow up studies explored methods for improving them further. XLNet \citep{yang2019xlnet} adds autoregressive capabilities to BERT, improving its quality, though at the cost of additional compute requirements. RoBERTa \citep{liu2019roberta} modifies the training procedure of BERT and provides pre-training methods that significantly improve its performance. Two notable papers exploring the architecture design of the BERT are following. \citet{michel2019sixteen} examines the importance of attention heads in BERT architecture, highlighting scenarios where attention heads may be pruned. The main objective of the paper is to provide techniques for pruning attention head, and as such the amount of experiments performed on BERT is limited to a single task (MNLI). ALBERT \citep{lan2019albert} proposes two methods for reducing the number of parameters in BERT. The first is via parameter sharing across layers, and the second is by factorizing the embedding layers. We note (this was mentioned in the conclusion section of the paper) that while these methods are efficient in reducing the number of parameters used by the model, they do not help in reducing its latency. 

These studies provide some advancement towards a more efficient architecture design for BERT but leave much to be explored. In this paper we take a broader approach examining multiple design choices. We parameterize each layer of BERT by five different dimensions, as opposed to \citet{devlin2018bert} that parameterizes a layer with two dimensions and suggests a fixed value for the remaining three. We then (pre-)train multiple variants of BERT with different values chosen for these dimensions by applying pruning-based architecture search technique that jointly optimizes the architecture of the model with the objective of minimizing both the pre-training loss and the number of model parameters. Our experiments result in the following findings:
%
%
\begin{itemize}
    \item The ratio of the architecture design dimensions within a BERT encoder layer can be modified to obtain a layer with better performance. Transformer design dimensions suggested in \citet{vaswani2017attention} are sub-optimal.  
    \vspace{-0.8em}
    \item When we aim to obtain a computationally lighter model, using a `tall and narrow' architecture provides better performance than a `wide and shallow' architecture.
    \vspace{-0.8em}
    \item The fully-connected component applied to each token separately plays a much more significant role in the top layers as compared to the bottom layers. 
\end{itemize}

\section{Background}
Following BERT's notations, we use $\ell$ to denote the number of encoder layers (i.e. Transformer blocks), $h$ to denote the hidden size, and $a$ to denote the number of self attention heads. The BERT paper \citep{devlin2018bert} primarily reports results on two models: {$\rm BERT_{BASE}$} $(\ell=12, h = 768, a = 12)$ and {$\rm BERT_{LARGE}$} $(\ell=24, h = 1024, a = 16)$. BERT base has $108$M parameters and BERT large has $340$M parameters. Though BERT large achieves higher accuracy than BERT base, due to its prohibitively large size it finds limited use in practice. Since BERT base achieves higher accuracy compared to previous state-of-the-art models- Pre-OpenAI SOTA, BiLSTM+ELMo+Attn and OpenAI GPT- on most of the benchmark datasets, it is widely used in practice. BERT base and OpenAI GPT have the same number of model parameters. 

Given its broad adoption for NLP tasks, an immediate question is: can we reduce the size of BERT base without incurring any significant loss in accuracy? The BERT paper \citep{devlin2018bert} provides an ablation study, Table \ref{tab:bert_ablation}, over the number of model parameters by varying the number of layers $\ell$, the hidden size $h$, and the number of attention heads $a$. 
\begin{table}
\begin{center}
\begin{tabular}{ccccccc}
\hline
\multicolumn{4}{c}{Design dimensions} & \multicolumn{3}{c}{Dev Set Accuracy}\\
\hline
$\#\ell$ & $\#h$ & $\#a$ & $\#$M &  MNLI & MRPC & SST-2\\
3 & 768 & 12 & $45$ & 77.9 & 79.8 & 88.4 \\
6 & 768 & 3 &  $55$ & 80.6 & 82.2 & 90.7 \\
6 & 768 & 12 & $66$ & 81.9 & 84.8 & 91.3 \\
\multicolumn{7}{l}{BERT base}\\
12 & 768 & 12 & $108$ & 84.4 & 86.7 & 92.9\\
\hline
\end{tabular}
\end{center}
\caption{\label{tab:bert_ablation} Ablation study over BERT model size, Table $6$ in \citet{devlin2018bert}. $\#$M denotes number of model parameters in millions.}
\end{table}
It can be observed that the accuracy decreases drastically when the number of encoder layers $\ell$ is reduced, and also when the number of attention heads is reduced. We ask the following question: are there any other design dimensions that can be reduced without incurring huge loss in accuracy?

As noted above, the three primary design dimensions of the BERT architecture are the number of encoder layers $\ell$, the hidden size $h$, and the number of attention heads $a$. BERT's Transformer encoder layers are based on the original Transformer implementation described in \citet{vaswani2017attention}. \citet{vaswani2017attention} fixed dimension of key, query, and value in multi-head attention, and filter dimension in feed-forward networks as a function of the hidden size and the number of attention heads. However, these are variable design dimensions and can be optimized. Moreover, BERT architecture uses the same number of attention heads for all the encoder layers and hence all the layers are identical. In this work, we jointly optimize all these design dimensions of BERT architecture while allowing each encoder layer to have different design dimensions. 

In order to explore the parameter space efficiently we chose to optimize the design dimensions in a pruning framework rather than launching a pre-training job for each of these choices. This allows a speedup of several orders of magnitude that is crucial in order to obtain meaningful conclusions. 
We parameterize the different dimensions one can modify and jointly optimize them with a mixed target of both accuracy and parameter reduction. We look at how the accuracy of BERT evolves on various downstream datasets like GLUE, SQuAD v1.1, and SQuAD v2.0 when we reduce the model size
via an optimization procedure.


\section{Related works}
There is a vast literature on pruning trained neural networks. Starting with the classical works \citet{lecun1990optimal,hassibi1993second}
in the early $90$'s to the recent works \citet{han2015deep}, pruning deep neural networks has received a lot of attention. There have been two orthogonal approaches in pruning networks: structured pruning \citep{li2016pruning, molchanov2016pruning} and unstructured pruning \citep{anwar2017structured}. Structured pruning gives smaller architecture whereas unstructured pruning gives sparse model parameters. 
In natural language processing, \citet{murray2015auto} explored structured pruning in feed-forward language models. \citet{see2016compression} and \citet{kim2016sequence} provided pruning approaches for machine translation. 
A closely related line of work is Neural Architecture Search (NAS). 
It aims to efficiently search the space of architectures \citep{pham2018efficient, liu2018darts,singh2019darc}. Quantization is another technique to reduce the model size. This is done by quantizing the model parameters to binary \citep{rastegari2016xnor, hubara2017quantized}, ternary \citep{zhu2016trained}, or $4$ or $8$ bits per parameter \citep{han2015deep}. 

Recently published DistilBERT \citep{sanh2019distilbert} shows that a BERT model with fewer number of layers can be efficiently pre-trained using knowledge distillation to give much higher accuracy as compared to the same model pre-trained in a regular way. We note that the distillation technique is complimentary to our work and our schuBERTs can be pre-trained using distillation to boost their accuracy.
The ablation study in Table \ref{tab:bert_ablation}, BERT \citep{devlin2018bert}, and the above explained works \citep{michel2019sixteen, lan2019albert} look at the problem of reducing the BERT model size by reducing one or the other design dimensions - number of encoder layers, hidden size, number of attention heads, and embedding size - in isolation and in a sub-optimal way. In this work, we address this problem comprehensively.


\section{The Elements of BERT}
In this section, we present detailed architecture of the original BERT model and explain which design dimensions of it can be optimized. 
\begin{figure}[h]
\centering
\includegraphics[scale=0.3]{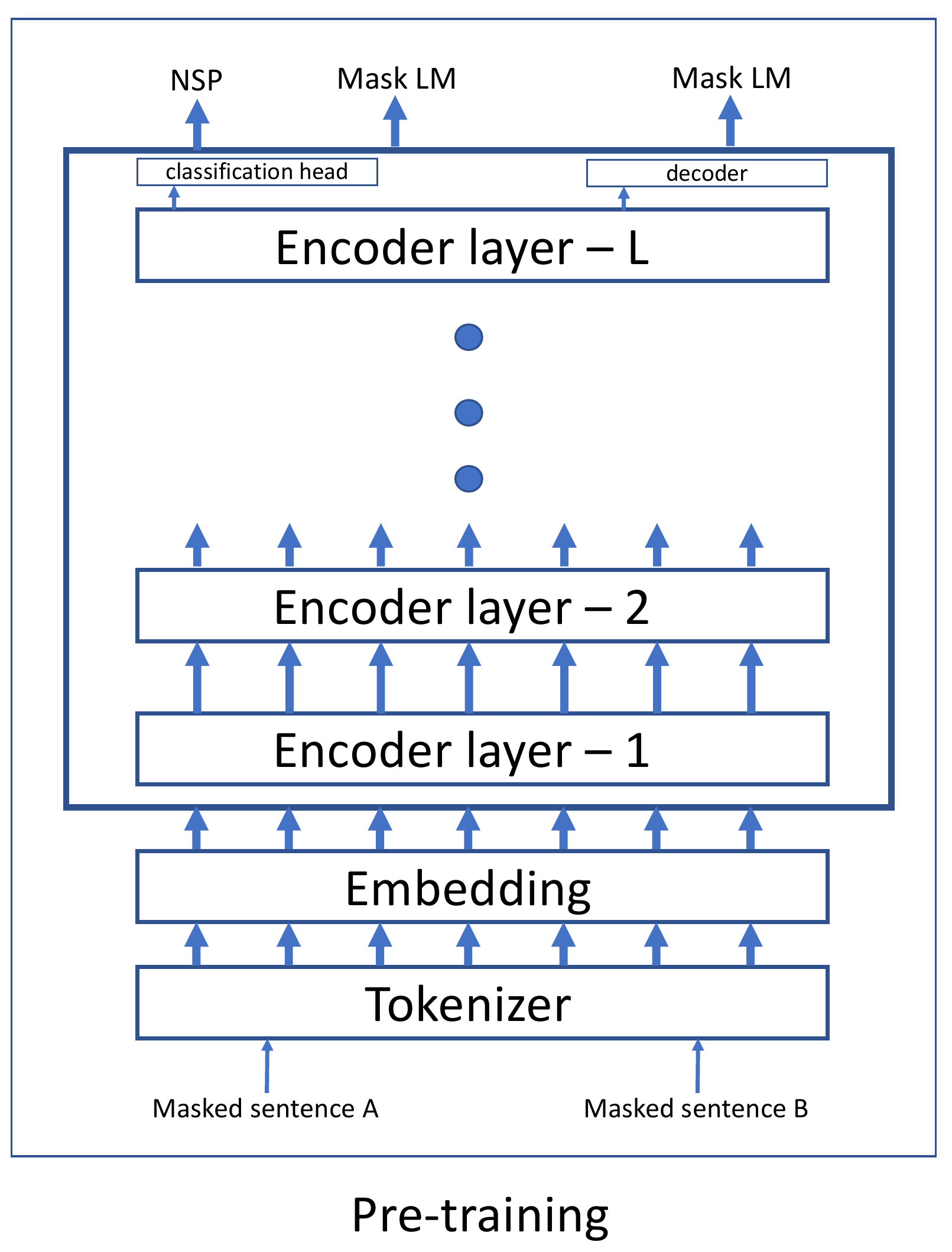}
\caption{BERT pre-training}
\label{fig:bert1}
\end{figure}
Figure \ref{fig:bert1} shows BERT pre-training architecture. First, the tokenized inputs are embedded into a vector of dimension $h$ through an embedding layer $E$. The embedded inputs pass through a sequence of encoder layers $1$ to $\ell$. Each encoder layer is identical in its architecture. The output of the last encoder layer is decoded using the same embedding layer $E$ and softmax cross-entropy loss is computed on the masked tokens. A special token CLS from the last encoder layer is used to compute next-sentence-prediction (NSP) loss. For further details of the loss corresponding to masked tokens and the NSP loss, we refer the readers to the BERT paper \citep{devlin2018bert}. 

We follow BERT notation conventions and denote the number of encoder layers as $\ell$, the hidden size as $h$, and the number of attention heads as $a$. Following the original Transformer implementation described in \citet{vaswani2017attention} BERT sets key-query dimension for multi-head attention $k$ to $h/a$. Following the same Transformer implementation it sets value dimension for multi-head attention $v$ equal to $k$, and feed-forward filter size $f$ equal to $4h$. In total, there are three design dimensions in BERT- $\ell$, $h$ and $a$, they are listed in Table \ref{tab:elem1}. For BERT base, the number of encoder layers $\ell$ is set to $12$, the hidden size $h$ is set to $768$, and the number of attention heads $a$ is set to $12$. The other three dimensions $f, k, v$ are function of $h$ and $a$. Further, each encoder layer of BERT is identical and uses same value of $a, f, k, v$.
\begin{table}
\begin{center}
\begin{tabular}{lcc}
BERT-base && \\
\hline
number of encoder layers & $\ell$ & $12$ \\
hidden size & $h$ & $768$ \\
number of self-attention heads & $a$ & $12$\\
feed forward dimension & $f$ & $4h$\\
key-query dimension for attention & $k$ & $h/a$\\
value dimension for attention & $v$ & $h/a$\\
\end{tabular}
\end{center}
\caption{\label{tab:elem1} Elements of BERT}
\end{table}

First of all, BERT has no architectural constraint that requires all the encoder layers to be identical. This aspect of design can be optimized and in full-generality it might result in highly non-identical layers. This implies that a generalized BERT will have $a_1, a_2, \cdots, a_\ell$ number of heads, $f_1, f_2, \cdots, f_\ell$ filter sizes in the feed forward networks, $k_{1}, k_{2}, \cdots, k_{\ell}$ key sizes and $v_{1}, v_{2}, \cdots, v_{\ell}$ value sizes in the attention heads, in the layers $1, 2, \cdots, \ell$ respectively.  Table \ref{tab:elem2} lists all the design dimensions of BERT that can be optimized without changing the architecture. Note that we abuse the term {\em architecture} to refer to the entire BERT network and the layer operations except sizes of the parameter matrices. In this work, our goal is to optimize (by pruning) all these dimensions to maximize accuracy for a given size of the model. We refer the BERT with optimized dimensions as schuBERT- Size Constricted Hidden Unit BERT.
\begin{table}
\begin{center}
\begin{tabular}{ll}
schuBERT \\
\hline
$\ell$ & $\ell$ \\
$h$ & $h$ \\
$a$ & $a_1, a_2, \cdots, a_{\ell}$ \\
$f$ & $f_1, f_2, \cdots, f_{\ell}$ \\
$k$ & $k_{1}, k_{2}, \cdots, k_{\ell}$ \\
$v$ & $v_{1}, v_{2}, \cdots, v_{\ell}$
\end{tabular}
\end{center}
\caption{\label{tab:elem2} Elements of schuBERT}
\end{table}

Now, we show which parameter matrices are tied with each of these design dimensions. Each design dimension is tied with more than one parameter matrix. This is explained by providing a detail view of an encoder cell of the BERT. 

\begin{figure}[h]
\centering
\includegraphics[scale=0.2]{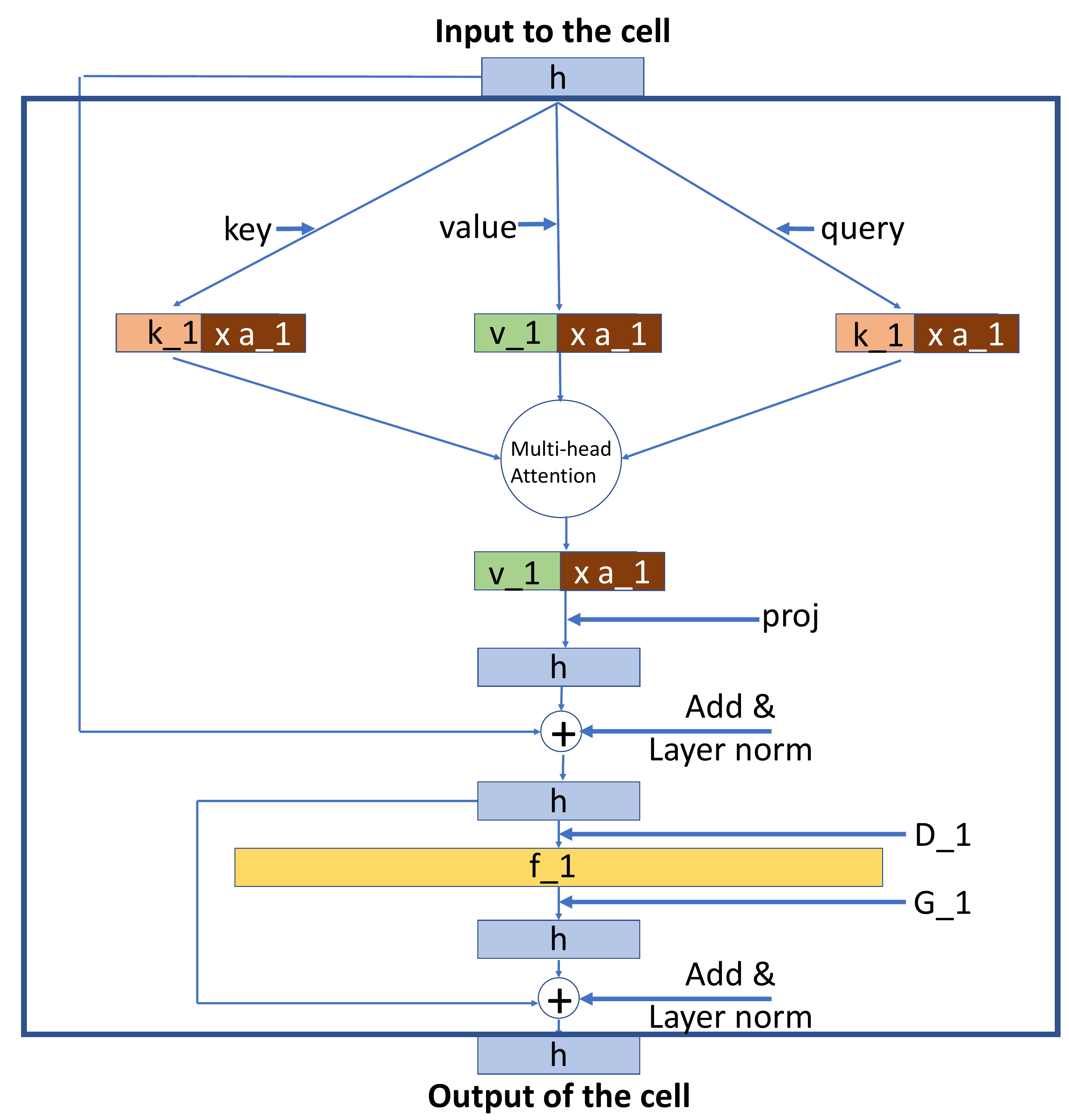}
\caption{An encoder layer of schuBERT}
\label{fig:bert2}
\end{figure}

Figure \ref{fig:bert2} shows architecture of an encoder layer of BERT. The notations in the figure have subscript $1$ that represent first encoder layer. Input to an encoder layer is the hidden representation of a token which is of dimension $h$. Input first goes through a multi-head attention cell. Note that multi-head attention cell processes hidden representation of all the tokens in a combined way. For simplicity, in Figure \ref{fig:bert2} we have shown only one hidden representation.

The multi-head attention cell consists of three parameter tensors, namely - {\em key} $K_1$, {\em query} $Q_1$ and {\em value} $V_1$. $K_1$ is of size $k_1\times a_1 \times h$. Key vector for each head of the attention is of dimension $k_1$ and $a_1$ represents the number of heads. Hidden representation of dimension $h$ is projected on the {\em key} tensor $K_1$ to get $a_1$ key vectors each of dimension $k_1$. Similarly the {\em query} tensor $Q_1$ is used to get $a_1$ query vectors each of dimension $k_1$ for $a_1$ heads of the multi-head attention cell. The {\em value} tensor $V_1$ is of dimension $v_1 \times a_1 \times h$. The hidden representation is projected on the {\em value} tensor $V_1$ to get $a_1$ value vectors each of dimension $v_1$. Note that $k_1$ and $v_1$ can be different. The inner product of key and query vectors after passing through softmax layer give weights for combining value vectors. For details of multi-head attention cell we refer the readers to \citet{vaswani2017attention}. In nutshell, using three parameter tensors- $K_1, Q_1, V_1$, a multi-head attention cell transforms hidden representation of size $h$ to a vector of dimension $(v_1\times a_1)$. This vector is projected back to the same dimension $h$ through a {\em proj} matrix $P_1$. Which is then added element-wise to the hidden representation that was input to the encoder cell and layer norm is applied on the addition. The output is passed sequentially through two fully-connected layers namely $D_1$ and $G_1$.  $D_1$ consists of a parameter matrix of dimension $f_1 \times h$ and $G_1$ consists of a parameter matrix of dimension $h \times f_1$. The output of $G_1$ is added element-wise to the input of $D_1$ and layer norm is applied to it. This is the output of the encoder cell and is input to the next encoder cell. 

The color coding in Figure \ref{fig:bert2} shows which vectors need to be of the same dimension. The hidden representation size $h$ needs to be same throughout all the encoder layers. In a multi-head attention cell, in each head key and query vectors must have the same dimension. Therefore, {\em key} and {\em query} tensors, $K_1, Q_1$ must be of the same size $k_1\times a_1 \times h$. The value vector can be of different dimension $v_1$. Therefore the {\em value} tensor $V_1$ should be of dimension $v_1\times a_1 \times h$.  Further, the filter size $f_1$ in the two fully-connected layers $D_1, G_1$ is a variable and can take any integral value. 

Keeping aligned with the BERT and the subsequent improvements such as XLNet \citep{yang2019xlnet} and RoBERTa \citep{liu2019roberta}, we set the WordPiece embedding size $e$ equal to the hidden layer size $h$, i.e. $e \equiv h$. However, factorization of the embedding matrix can be incorporated as demonstrated in ALBERT \citep{lan2019albert}.

\section{Optimization Method}
We optimize BERT design dimensions listed in Table \ref{tab:elem2} by pruning the original BERT base architecture. All the design dimensions are upper bounded by their original value in the BERT base as given in the Table \ref{tab:elem1}. Since we keep the architecture same, that is we do not remove any layer, the design dimensions are lower bounded by one. 

For each design dimension that we seek to optimize, we introduce a prune-parameter vector $\alpha$ of size equal to the original dimension. We take pre-trained original BERT base network, and multiply all the parameter tensors/matrices that are associated with the particular design dimension with the corresponding prune-parameter vector. For example, filter size of the feed-forward layer in the first encoder layer is $f_1 = 3072$. To optimize $f_1$, we introduce a prune-parameter vector $\alpha_{f_1} \in \mathbf{R}^{3072}$ and initialize it with all ones. In the original BERT base, the two parameter matrices $D_1$ and  $G_1$ are associated with the design dimension $f_1$. 
We replace $D_1$ by ${\rm diag}(\alpha_{f_1}) \cdot D_1$ and 
$G_1$ by $G_1 \cdot {\rm diag}(\alpha_{f_1})$ in the BERT pre-trained model.

Table \ref{tab:prune1} lists all the prune parameters. Table \ref{tab:prune2} lists all the parameter tensors/matrices for which design dimensions are optimized by multiplying prunable parameters on all the sides. {\em key} and {\em query} tensors $K_i, Q_i$ for $i \in \{1,2,\cdots,\ell\}$ are multiplied on all the three sides with prunable parameters corresponding to key-vector, number of attention heads, and hidden size. Similarly multiplications are performed on {\em value} tensor $V_i$ with a different value-vector prunable parameter. {\em proj} tensor has same multiplication as {\em value} tensor. The two feed-forward matrices $D_i, G_i$ have same multiplications. We denote the so obtained prunable tensors with tilde on their top. Note that we do not have prune parameters for pruning encoder layers. We find the optimal number of encoder layers $\ell$ by running experiments for different values of $\ell$. 

Our approach is to optimally find which individual elements of prunable parameters $\{\alpha_h,\{\alpha_{a_i}, \alpha_{v_i}, \alpha_{k_i},  \alpha_{f_i}\}_{i \in [\ell]} \}$ can be set to zero while incurring minimal increase in the pre-training loss. After we have sparse prunable parameter vectors, we remove the corresponding rows/columns from the BERT parameter matrices $\{K_i, Q_i, V_i, P_i, D_i, G_i\}_{i \in [\ell]}$, and get a smaller/faster BERT model. Below we explain the algorithm to find the sparse prunable parameters. 
\begin{table}
\begin{center}
\begin{tabular}{l}
$\alpha_h \in \mathbf{R}^{h}$ \\
$\{\alpha_{f_i} \in \mathbf{R}^{f} \}_{i=1,2,\cdots,\ell}$\\
$\{\alpha_{a_i} \in \mathbf{R}^{a}\}_{i=1,2,\cdots,\ell}$\\
$\{\alpha_{k_i} \in \mathbf{R}^{k}\}_{i=1,2,\cdots,\ell}$\\
$\{\alpha_{v_i} \in \mathbf{R}^{v} \}_{i=1,2,\cdots,\ell}$
\end{tabular}
\end{center}
\caption{\label{tab:prune1} Prunable parameters.}
\end{table}
\begin{table}
\begin{center}
\begin{tabular}{l}
$K_i \in \mathbf{R}^{k\times a \times h} \rightarrow K_i[\mathcal{D}(\alpha_{k_i}) \mathcal{D}(\alpha_{a_i}) \mathcal{D}(\alpha_{h})] \equiv \widetilde{K_i}$ \\
$Q_i \in \mathbf{R}^{k\times a \times h} \rightarrow Q_i[\mathcal{D}(\alpha_{k_i}) \mathcal{D}(\alpha_{a_i})  \mathcal{D}(\alpha_{h})] \equiv \widetilde{Q_i}$ \\
$V_i \in \mathbf{R}^{v\times a \times h} \rightarrow V_i[ \mathcal{D}(\alpha_{v_i}) \mathcal{D}(\alpha_{a_i})   \mathcal{D}(\alpha_{h})] \equiv \widetilde{V_i}$ \\
$P_i \in \mathbf{R}^{h \times v\times a} \rightarrow P_i[ \mathcal{D}(\alpha_{h})  \mathcal{D}(\alpha_{v_i}) \mathcal{D}(\alpha_{a_i})] \equiv \widetilde{P_i}$ \\
$D_i \in \mathbf{R}^{f \times h} \rightarrow  D_i[\mathcal{D}(\alpha_{f_i}) \mathcal{D}(\alpha_{h})] \equiv \widetilde{D_i}$ \\
$G_i \in \mathbf{R}^{h \times f} \rightarrow G_i [\mathcal{D}(\alpha_{h}) \mathcal{D}(\alpha_{f_i})] \equiv \widetilde{G_i}$ \\
\end{tabular}
\end{center}
\caption{\label{tab:prune2} Prunable BERT parameter matrices/tensors.}
\end{table}

We start with the pre-trained BERT base trained on BooksCorpus ($800$M words) and English Wikipedia ($2500$M words) following the BERT pre-training procedure given in \citet{devlin2018bert}. 
Particularly, we minimize the loss given in Equation \eqref{eq:org_loss} to learn the optimal parameter tensors $\{K_i, Q_i, V_i, P_i, D_i, G_i\}_{i \in [\ell]}$ and the embedding matrix $E$. Next, we introduce the prunable parameters given in Table \ref{tab:prune1} and initialize them with all ones. We create prunable BERT parameter matrices by multiplying the prunable parameters to the learned BERT parameter matrices, as given in Table \ref{tab:prune2}. Then, we optimize the prunable parameters $\alpha$'s while fixing the learned parameters matrices as given in Equation \ref{eq:prune_loss}. In addition to the MLM and NSP loss, we add sparsity inducing loss on the prunable parameters with a regularization coefficient $\gamma$. It is well known that $\ell_1$ penalty induces sparsity in the parameters. Further, since our goal is to minimize the number of parameters, to account for the fact that each element of prune parameters $\alpha$ when set to zero reduces different number of BERT parameters, we multiply the $\ell_1$ loss terms with the cost terms $\beta$'s. For example, $\beta_{a_i}$ is proportional to the number of model parameters that will be removed when an element of the prune parameter $\alpha_{a_i}$ is set to zero. It is critical to incorporate $\beta$'s. Their values are significantly different from each other. The $\beta$ values are $1.0, 0.73, 0.093, 0.093, 0.0078$ for $a,h,k,v$ and $f$ respectively. 

After training the prunable BERT model for a fixed number of steps, we truncate the smallest prune parameters to zero, and remove the corresponding rows/columns from the BERT parameter matrices $\{K_i, Q_i, V_i, P_i, D_i, G_i\}_{i \in [\ell]}$. Then we fine-tune the so obtained smaller schuBERT model. 

Algorithm \ref{algo:algo1} summarizes our approach. If we want to reduce the number of parameters by a fraction $\eta$, we do so in $T$ steps. In each step, we prune $\eta/T$ fraction of parameters, and at the end of the step we fine-tune the network and repeat these steps $T$ times. 
Though we have explained the algorithm in terms of $\ell_1$ penalty on the prunable parameters, in our experiments we tried alternative sparsity inducing penalties as well- $\ell_0$ regularization, and proximal gradient descent on prunable parameters. 
\begin{align}
& {\arg\min}_{\{E,\{K_i, Q_i, V_i, P_i, D_i, G_i\}_{i \in [\ell]} \}} \nonumber\\
    & \mathcal{L_{\rm MLM + NSP}}(E,\{K_i, Q_i, V_i, P_i, D_i, G_i\}_{i \in [\ell]})\,.      \label{eq:org_loss} \\
& {\arg\min}_{\{\alpha_h,\{\alpha_{a_i}, \alpha_{v_i}, \alpha_{k_i},  \alpha_{f_i}\}_{i \in [\ell]} \}} \nonumber\\
    & \mathcal{L_{\rm MLM + NSP}}(E,\{\widetilde{K_i}, \widetilde{Q_i}, \widetilde{V_i}, 
    \widetilde{P_i}, \widetilde{D_i}, \widetilde{G_i}\}_{i \in [\ell]}) \nonumber\\
        & + \gamma\{\beta_h\|\alpha_h\|\} + \gamma\sum_{i=1}^{\ell}\{\beta_{a_i}\|\alpha_{a_i}\|
         + \beta_{v_i}\|\alpha_{v_i}\| \nonumber\\
         & \quad\quad \quad+ \beta_{k_i}\|\alpha_{k_i}\| 
         + \beta_{f_i}\|\alpha_{f_i}\| \} \,.
    \label{eq:prune_loss}
\end{align}
\renewcommand{\algorithmicrequire}{\textbf{Input:}}
\renewcommand{\algorithmicensure}{\textbf{Output:}}
\begin{center}
\begin{algorithm}
\caption{Pruning Transformers}
\begin{algorithmic} \label{algo:algo1}
    \REQUIRE A Transformer model, minimization objective (FLOPs/Params/Latency), target fraction $\eta$, number of iterations $T$.
	\ENSURE A optimally pruned Transformer model.
    	\STATE \hspace{-1em} {\textbf{pre-training: }Train the network using loss Equation \eqref{eq:org_loss}.}
	\STATE \hspace{-1em}\textbf{Repeat $T$ times:} 
	\STATE $\bullet$ Initialize prunable parameters $\alpha_h, \alpha_{a_i}, \alpha_{k_i}, \alpha_{v_i}, \alpha_{f_i} \rightarrow \mathbf{1}$
	\STATE $\bullet$  Multiply prunable parameters with network parameter
	\STATE $K_i, Q_i, V_i, P_i, D_i, G_i \rightarrow \widetilde{K_i}, \widetilde{Q_i},
	\widetilde{V_i}, \widetilde{P_i}, \widetilde{D_i}, \widetilde{G_i}$
	\STATE $\bullet$ Train the network using loss Equation \eqref{eq:prune_loss}
	\STATE $\bullet$  Set the $\zeta$ smallest prunable parameters to zero to achieve $\eta/T$ reduction in the target objective value
	\STATE $\bullet$ Offset zero and non-zero prunable parameters into model parameters
	\STATE $\widetilde{K_i}, \widetilde{Q_i},
	\widetilde{V_i}, \widetilde{P_i}, \widetilde{D_i}, \widetilde{G_i} \rightarrow   K_i, Q_i, V_i, P_i, D_i, G_i$	
	\STATE $\bullet$ Create smaller model parameter tensors by removing all-zero rows/columns
	\STATE $K_i, Q_i, V_i, P_i, D_i, G_i \rightarrow \widehat{K_i}, \widehat{Q_i},
	\widehat{V_i}, \widehat{P_i}, \widehat{D_i}, \widehat{G_i}$
	\STATE $\bullet$ Finetune the model using loss Equation \eqref{eq:org_loss}
\end{algorithmic}
\end{algorithm}
\end{center}
\section{Experimental Results}
\begin{table*}[h]
\centering
\begin{tabular}{lcccccc|c}
model & SQuAD v1.1 & SQuAD v2.0 & MNLI & MRPC & SST-2 & RTE & Avg\\
\hline 
BERT-base ($108$M) & $90.2/83.3$ &$80.4/77.6$& $84.1$ & $87.8$ & $92.1$ & $71.4$ & $84.3$\\
\multicolumn{8}{c}{\# parameters = $99$M}\\
\hline
schuBERT-all & $\pmb{89.8/83.0}$ &$\pmb{80.1/77.6}$& $\pmb{83.9}$ & $\pmb{87.5}$ & $\pmb{92.4}$ & $\pmb{71.1}$ & $\pmb{84.1}$\\
schuBERT-$f$ & $89.8/82.9$ &$79.6/77.3$& $83.5$ & $87.4$ & $91.6$ & $70.7$ & $83.8$\\
schuBERT-$h$ & $89.6/82.6$ &$79.9/77.5$& $83.7$ & $87.3$ & $91.5$ & $70.4$ & $83.7$\\
BERT-all uniform & $89.7/82.7$ &$79.8/77.3$& $83.7$ & $87.2$ & $92.0$ & $69.8$ & $83.7$\\
schuBERT-$a$ & $89.3/82.3$ &$79.1/77.4$& $83.3$ & $86.8$ &$91.1$ &$69.1$ & $83.1$\\
\end{tabular}
\caption{\label{tab:results_99}Accuracy results on SQuAD and GLUE datasets obtained by fine-tuning BERT and schuBERTs with total of $99$ million parameters.}
\end{table*}

\begin{table*}[h]
\centering
\begin{tabular}{l|cccccccccccc}
\hline
$\ell$ & $1$ & $2$ & $3$ & $4$ & $5$ & $6$& $7$ &$8$ & $9$ & $10$ & $11$ & $12$\\
\hline
$f = $ & $2022$ &  $2222$ &  $2344$ & $2478$ & $2576$ & $2530$ &  $2638$ & $2660$ & $2748$ & $2792$ & $2852$ & $2974$\\
$a = $ & $12$ & $12$ & $12$ & $12$ & $11$ & $12$ & $12$ & $12$ &  $12$ & $12$ & $12$ & $12$\\
$k= $ &$64$ & $64$ & $64$ & $64$ & $64$ & $64$ & $64$ & $64$ & $64$ & $64$ & $64$ & $64$\\
$v = $ & $54$ & $54$ & $46$ & $58$ & $52$ & $60$ &  $64$ & $64$ & $64$ & $64$ & $64$ & $62$\\
\hline
\multicolumn{12}{l}{number of encoder layers $\ell=12$, number of hidden units $h=768$}\\
\end{tabular}
\caption{\label{tab:arch_99} Design dimensions of schuBERT-all for $99$ million parameters.}
\end{table*}

\begin{table*}[h]
\centering
\begin{tabular}{lcccccc|c}
model & SQuAD v1.1 & SQuAD v2.0 & MNLI & MRPC & SST-2 & RTE & Avg\\
\hline 
BERT-base ($108$M) & $90.2/83.3$ &$80.4/77.6$& $84.1$ & $87.8$ & $92.1$ & $71.4$ & $84.3$\\
\multicolumn{8}{c}{\# parameters = $88$M}\\
\hline
BERT-$\ell$ & $88.4/80.9$ &$78.8/77.2$& $83.8$ & $85.6$ & $91.3$ & $68.2$ & $82.7$\\
schuBERT-all & $89.4/82.5$ &$79.8/77.1$& $\pmb{84.1}$ & $\pmb{87.6}$ & $\pmb{92.3}$ & $\pmb{69.7}$ & $\pmb{83.8}$\\
schuBERT-$f$ & $89.2/82.2$ &$79.5/77.5$& $83.7$ & $87.4$ & $92.2$ & $69.3$ & $83.6$\\
BERT-all uniform & $89.1/82.0$ &$79.6/77.6$& $83.7$ & $87.5$ & $91.7$ & $68.9$ & $83.4$\\
schuBERT-$h$ & $89.1/82.0$ &$79.4/77.3$& $83.6$ & $87.2$ & $91.5$ & $69.2$ & $83.3$\\
schuBERT-$a$ & $85.1/77.1$ &$74.1/72.4$& $82.2$ & $85.2$ & $90.9$ & $67.0$ & $80.8$\\
ALBERT-$e$ & $\pmb{89.9/82.9}$ &$\pmb{80.1/77.8}$& $82.9$ &$-$& $91.5$ &$-$ & $-$
\end{tabular}
\caption{\label{tab:results_88}Accuracy results on SQuAD and GLUE datasets obtained by fine-tuning BERT, ALBERT, and schuBERTs with total of $88$ million parameters.
}
\end{table*}

\begin{table*}[h]
\centering
\begin{tabular}{l|cccccccccccc}
\hline
$\ell$ & $1$ & $2$ & $3$ & $4$ & $5$ & $6$& $7$ &$8$ & $9$ & $10$ & $11$ & $12$\\
\hline
$f = $ & $1382$ & $1550$ & $1672$ & $1956$ & $2052$ & $2030$ & $2210$ & $2314$ & $2474$ & $2556$ & $2668$ & $2938$\\
$a = $ & $12$ & $12$ & $11$ & $12$ & $11$ & $12$ & $12$ & $12$ & $12$ & $12$ & $12$ & $12$ \\
$k= $  & $64$ & $64$ & $64$ & $64$ & $64$ & $64$ & $64$ & $64$ & $64$ & $64$ & $64$ & $64$\\
$v = $ & $46$ & $48$ & $42$ & $52$ & $46$ & $54$ & $64$ & $62$ & $64$ & $64$ & $64$ & $40$\\
\hline
\multicolumn{12}{l}{number of encoder layers $\ell=12$, number of hidden units $h=756$}\\
\end{tabular}
\caption{\label{tab:arch_88} Design dimensions of schuBERT-all for $88$ million parameters.}
\end{table*}
\begin{table*}[h]
\centering
\begin{tabular}{lcccccc|c}
\multicolumn{8}{c}{\# parameters = $77$M}\\
\hline
model & SQuAD v1.1 & SQuAD v2.0 & MNLI & MRPC & SST-2 & RTE & Avg\\
\hline
schuBERT-$h$ & $\pmb{88.8/81.6}$ &$\pmb{78.6/76.3}$& $\pmb{84.0}$ & $\pmb{87.2}$ & $91.5$ & $\pmb{68.9}$ & $\pmb{83.2}$\\
BERT-all uniform & $88.8/81.6$ &$78.4/76.0$& $83.7$ & $86.6$ & $91.9$ & $68.9$ & $83.1$\\
schuBERT-$f$ & $88.8/81.4$ &$78.8/76.1$& $83.2$ & $86.5$ & $\pmb{92.2}$ & $67.7$ & $82.9$\\
schuBERT-all & $88.8/81.6$ &$78.6/76.2$& $83.8$ & $86.6$ & $92.2$ & $66.4$ & $82.7$\\
schuBERT-$a$ & $82.6/74.2$ &$73.1/68.9$& $82.0$ & $84.9$ & $89.6$ & $66.4$ & $79.8$\\
\end{tabular}
\caption{\label{tab:results_77}Accuracy results on SQuAD and GLUE datasets obtained by fine-tuning BERT and schuBERTs with total of $77$ million parameters.
}
\end{table*}

\begin{table*}[h]
\centering
\begin{tabular}{lcccccc|c}
\multicolumn{8}{c}{\# parameters = $66$M}\\
\hline
model & SQuAD v1.1 & SQuAD v2.0 & MNLI & MRPC & SST-2 & RTE & Avg\\
\hline
BERT-$\ell$ & $85.3/77.1$ &$75.3/72.5$& $82.3$ & $84.4$ & $91.1$ & $67.6$ & $81.0$\\
schuBERT-$h$ & $\pmb{88.1/80.7}$ &$\pmb{78.4/74.7}$& $\pmb{83.8}$ & $\pmb{86.7}$ & $\pmb{91.7}$ & $\pmb{68.5}$ & $\pmb{82.9}$\\
schuBERT-all & $88.0/80.7$ &$78.2/74.5$& $83.2$ & $87.2$ & $91.3$ & $67.8$ & $82.6$\\
BERT-all uniform & $87.7/80.3$ &$77.8/74.0$& $83.6$ & $86.2$ & $91.3$ & $68.1$ & $82.4$\\
schuBERT-$f$ & $87.6/80.0$ &$77.6/74.1$& $83.0$ & $86.8$ & $90.6$ & $68.1$ & $82.3$
\end{tabular}
\caption{\label{tab:results_66}Accuracy results on SQuAD and GLUE datasets obtained by fine-tuning BERT and schuBERTs with total of $66$ million parameters.
}
\end{table*}

\begin{table*}
\centering
\begin{tabular}{lcccccc|c}
\multicolumn{8}{c}{\# parameters = $55$M}\\
\hline
model & SQuAD v1.1 & SQuAD v2.0 & MNLI & MRPC & SST-2 & RTE & Avg\\
\hline
schuBERT-$h$ & $\pmb{87.6/80.3}$ &$\pmb{77.4/74.6}$& $\pmb{83.5}$ & $\pmb{86.3}$ & $\pmb{90.9}$ & $66.7$ & $\pmb{82.1}$\\
schuBERT-all & $86.8/79.3$ &$76.6/73.5$& $83.4$ & $86.3$ & $90.9$ & $66.8$ & $81.8$\\
BERT-all uniform & $86.2/78.5$ &$76.9/72.2$& $83.2$ & $84.0$ & $90.5$ & $67.1$ & $81.3$\\
schuBERT-$f$ & $85.8/77.5$ &$75.8/71.8$& $81.8$ & $84.4$ & $90.2$ & $\pmb{67.3}$ & $80.9$\\
\end{tabular}
\caption{\label{tab:results_55}Accuracy results on SQuAD and GLUE datasets obtained by fine-tuning BERT and schuBERTs with total of $55$ million parameters.
}
\end{table*}

\begin{table*}[h]
\centering
\begin{tabular}{lcccccc|c}
\multicolumn{8}{c}{\# parameters = $43$M}\\
\hline
model & SQuAD v1.1 & SQuAD v2.0 & MNLI & MRPC & SST-2 & RTE & Avg\\
\hline
BERT-$\ell$ & $75.6/65.8$ &$65.9/57.8$& $78.5$ & $79.5$ & $87.3$ & $63.8$ & $75.1$\\
schuBERT-$h$ & $\pmb{86.7/79.0}$ &$\pmb{76.9/73.8}$& $\pmb{83.4}$ & $\pmb{84.8}$ & $\pmb{90.9}$ & $\pmb{67.3}$ & $\pmb{81.7}$\\
schuBERT-all & $86.0/77.9$ &$76.7/72.8$& $82.6$ & $84.2$ & $90.5$ & $66.2$ & $81.0$\\
BERT-all uniform & $85.0/77.2$ &$75.3/72.4$& $82.2$ & $83.4$ & $90.6$ & $67.2$ & $80.6$\\
schuBERT-$f$ & $84.2/75.5$ &$74.7/69.8$& $80.3$ & $77.1$ & $89.7$ & $58.7$ & $77.5$
\end{tabular}
\caption{\label{tab:results_43}Accuracy results on SQuAD and GLUE datasets obtained by fine-tuning BERT and schuBERTs with total of $43$ million parameters.
}
\end{table*}

\begin{table*}[h]
\centering
\begin{tabular}{c|cccccccc}
\hline
\# parameters & BERT-base & $99$M & $88$M & $77$M & $66$M & $55$M & $43$M & $33$M\\
\hline
$\ell =$ & $12$ & $12$ & $12$ & $12$ & $12$ & $12$ & $12$ & $12$\\
$h =$  & $768$ & $768$ & $756$ & $544$ & $466$ & $390$ & $304$ & $234$\\
$f(\min-\max) =$  & $3072$& $2022-2974$ & $1382-2938$ & $3072$ & $3072$ & $3072$ & $3072$ & $3072$\\
$a(\min-\max) =$  & $12$& $11-12$ & $11-12$ & $12$ & $12$ & $12$ & $12$ & $12$\\
$k(\min-\max) =$  & $64$& $64$ & $64$ & $64$ & $64$ & $64$ & $64$ & $64$\\
$v(\min-\max) =$  & $64$& $46-64$ & $40-64$ & $64$ & $64$ & $64$ & $64$ & $64$\\
\hline
\end{tabular}
\caption{\label{tab:schuBERTs} Best schuBERT architectures for different number of model parameters. BERT base has $108$M parameters.}
\end{table*}

In this section, we present our experimental results. We apply Algorithm \ref{algo:algo1} on BERT base. For pre-training BERT base we use MXNET based gluon-nlp repository that uses the hyper-parameters suggested in the original BERT paper. Besides pre-training, our algorithm has three hyper-parameters: regularization coefficient $\gamma$, learning rate for prunable parameters, and the number of steps for regularizing prune parameters- Equation \eqref{eq:prune_loss}. We run hyper-parameter optimization on these parameters to get the best results. For regularization loss \eqref{eq:prune_loss}, we use the same training data that we use for pre-training, BooksCorpus (800M words) and English Wikipedia (2,500M words). However, we run the regularization step for $1/1000$th steps as used for pre-training. We finet-une the pruned BERT by training for $1/20$th of the steps used for pre-training. 

We provide accuracy results for schuBERT on the following downstream tasks- question answering datasets- SQuAD v1.1, SQuAD v2.0; and GLUE datasets - MNLI, MRPC, SST-2 and RTE. For these downstream tasks, we use the fine-tuning hyper-parameters as suggested in the BERT paper. 

We create six schuBERTs by pruning one or all of the design dimensions. 
Accuracy of the downstream tasks on these schuBERTs are given in Tables \ref{tab:results_99}-\ref{tab:results_43}.
The BERT base has $108$ million parameters. The schuBERT sizes $88, 66, 43$ million are chosen to match the number of parameters in BERT with $\ell \in \{9, 6, 3\}$ layers. 

We use schuBERT-$x$ notation for $x \in \{h, f, a\}$ to denote a schuBERT obtained by only pruning $h$-hidden size, $f$-filter size of feed-forward, $a$-number of attention heads respectively. We use schuBERT-all to denote the case when all the design dimensions- $h, f, a, k, v$, except $\ell$ are pruned. 

We compare our results with original BERT base, and by varying its number of encoder layers $\ell \in \{12, 9, 6, 3\}$. We denote these results by BERT-$\ell$. Since ALBERT reduces parameters by factorizing the embedding matrix, we denote its results by ALBERT-$e$. ALBERT provided results only for $88$ million parameter model, not for any smaller models. 
Further, we also compare with the baseline case when all the design dimensions are pruned uniformly. We denote these results by BERT-all uniform. 

For $99$M model, Table \ref{tab:results_99}, schuBERT-all beats the baseline BERT-all uniform by $0.4\%$ higher average accuracy and performs better than schuBERT-$f/h/a$. Moreover, the loss in performance in comparison to BERT base with $108$ million parameters is only $0.2\%$. Table \ref{tab:arch_99} gives exact design dimensions for schuBERT-all with $99$ million parameters. We see that number of hidden units remain same as in BERT base, $h=768$. Parameter reduction primarily comes from feed-forward layers. Moreover, filter size of feed-forward layer - $f$ has a clear increasing pattern across the layers.  

For $88$M model, Table \ref{tab:results_88}, again schuBERT-all beats all the other models.  It gives $1.1\%$ higher average accuracy than BERT-$\ell$ with $9$ layers. ALBERT-$e$ performs better on SQuAD datasets, but performs significantly worse on MNLI and SST-2 datasets. Note ALBERT's approach is complementary to our approach and it can be incorporated into our schuBERTs.
schuBERT-$a$ performs significantly worse than schuBERT-all which implies that pruning only number of attention heads is highly sub-optimal, as is recently done in \citet{michel2019sixteen}. 
Table \ref{tab:arch_88} provides the exact design dimensions for schuBERT-all with $88$ million parameters. Similar to $99$M model, filter size of feed-forward layer - $f$ has a clear increasing pattern across the layers.  


For heavily pruned models - $77$M, $66$M, $55$M and $43$M models - accuracy results are shown in Table \ref{tab:results_77}, Table \ref{tab:results_66}, Table \ref{tab:results_55} and Table \ref{tab:results_43} respectively. In all these models 
schuBERT-$h$ beats all the other models. For $66$M model, schuBERT-$h$ gives $1.9\%$ higher average accuracy than BERT-$\ell$ with $6$ layers. For $43$M model, schuBERT-$h$ gives $6.6\%$ higher average accuracy than BERT-$\ell$ with $3$ layers. That is reducing the hidden units is way better than to reduce the number of layers to create a light BERT model. Ideally, we would expect schuBERT-all to perform better than schuBERT-$h$, but marginally worse performance of schuBERT-all can be attributed to the high complexity of pruning all the design dimensions together. 

Table \ref{tab:schuBERTs} provides best schuBERT architectures when the number of model parameters are restricted to different values. For smaller models, schuBERT-$h$ outperforms all other schuBERTs including schuBERT-all. Note that our schuBERT architectures are smaller in size as well as they yield lower latency. 

\section{schuBERT}
Based on the above described experimental results, we provide following insights on the design dimensions of schuBERT architecture. 

\begin{figure}
 \begin{center}
	\includegraphics[width=0.7\linewidth]{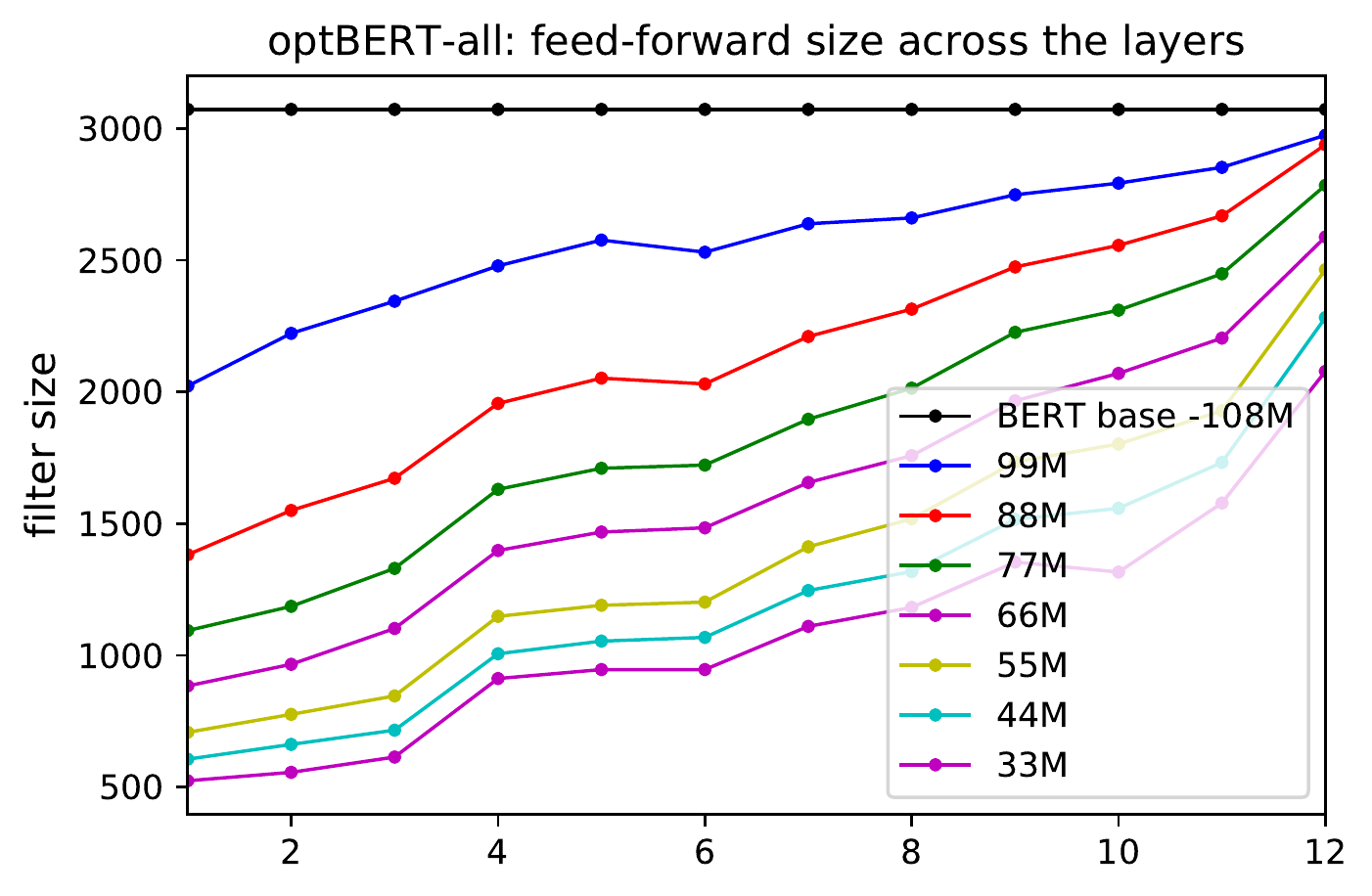}
	\caption{Feed-forward size across the encoder layers in schuBERT-all for various model sizes.    
	}
	\vspace{-1em}
	\label{fig:hidden}
\end{center}
\end{figure}

\textbf{Slanted Feed-forward Layer.} The fully-connected component applied to each token separately plays a much more significant role in the top layers as compared to the bottom layers. Figure \ref{fig:hidden} shows pattern of filter size of feed-forward layer across the encoder cells for various schuBERT-all models. In each of them, filter size follows an increasing pattern with min-max ratio ranging from $1.5$ to $4$, as opposed to same value across all the layers.

\textbf{Tall and Narrow BERT.} When we aim to obtain a computationally lighter model, using a `tall and narrow' architecture provides better performance than a `wide and shallow' architecture. Our results in Tables \ref{tab:results_88}, \ref{tab:results_66}, \ref{tab:results_43} demonstrate that schuBERT with $\ell=12$ encoder layers significantly outperforms BERT with $\ell \in \{9,6,3\}$ layers for the same number of parameters.

\textbf{Expansive Multi-head Attention.} The ratio of the design dimensions within a BERT encoder layer can be modified to obtain a better performing layer architecture. Transformer design dimensions suggested in \citep{vaswani2017attention} are sub-optimal. 

Following the original Transformer architecture described in \citep{vaswani2017attention}, BERT and other Transformer based models set key-query $k$ and value $v$ dimension for multi-head attention to $k=v=h/a$, where $h$ is the size of the hidden representation, and $a$ is the number of attention heads. Also, following the same architecture \citep{vaswani2017attention}, BERT sets feed-forward filter size $f=4h$. Although there is no restriction in using different output dimensions $k, v$ and filter size $f$, without changing the behaviour of the attention mechanism, we are not aware of any study questioning this `default value' of $k=v=h/a$ and $f=4h$. 

Our schuBERT architecture for various model sizes given in Table \ref{tab:schuBERTs}, show that for smaller models $k,v$ should be much larger than $h/a$. For $43$M schuBERT model $h/a = 25.3$ whereas $k=v=64$. Also, $f$ should be much larger than $4h$. For the same $43$M schuBERT model $4h=936$ whereas $f=3072$. Table \ref{tab:results_43} shows that $43$M schuBERT  ($\ell=12, h = 304, a= 12, k=v=64, f=3072$) significantly outperforms BERT-$\ell$ ($\ell=3, h = 768, a = 12, k=v=h/a, f=4h$).

\newpage


\clearpage
\newpage
\bibliography{bert}
\bibliographystyle{acl_natbib}

\appendix
\section{Appendix}
\label{sec:appendix}

\begin{figure}[h]
 \begin{center}
	\includegraphics[width=0.9\linewidth]{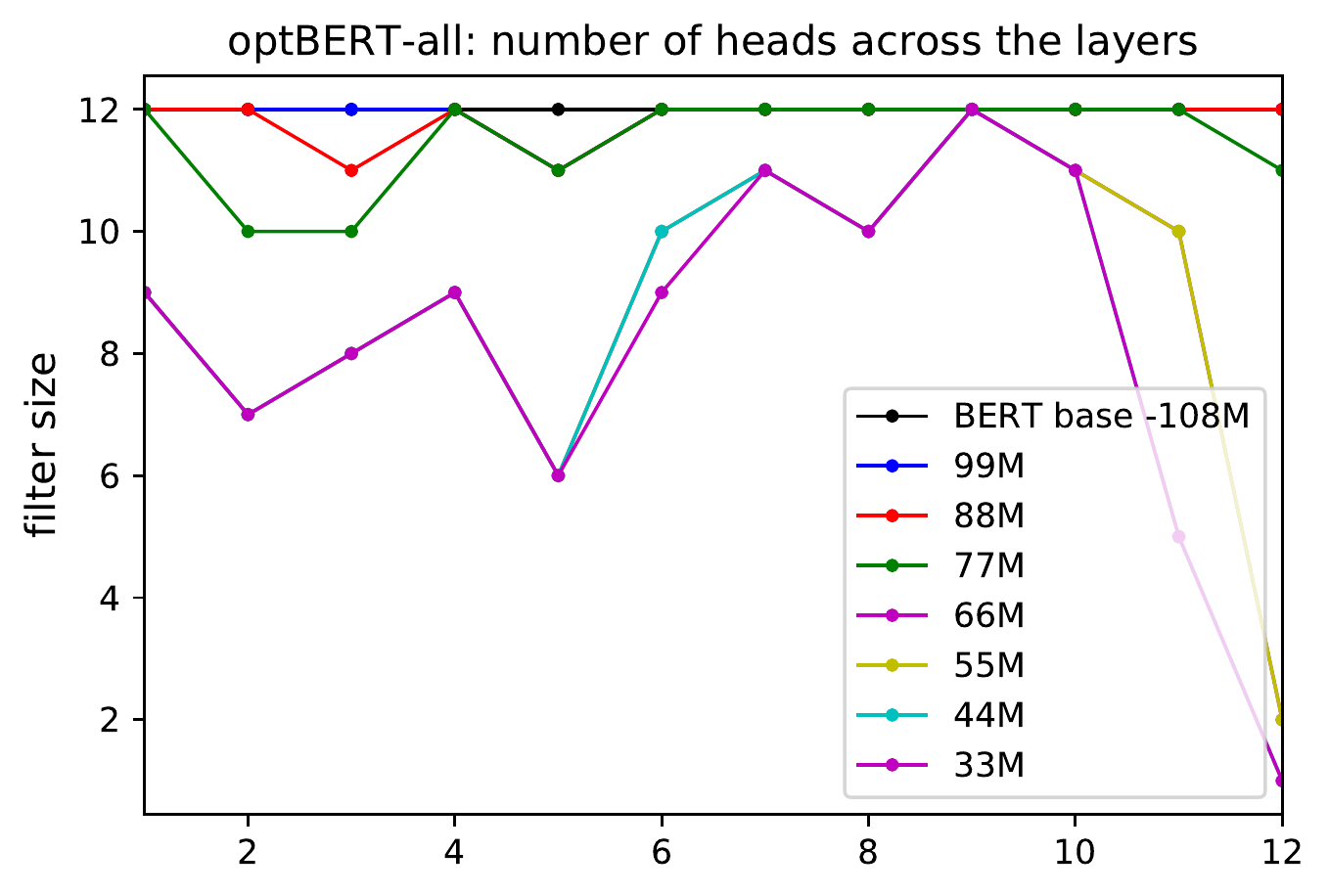}
	\caption{Number of multi-attention heads across the encoder layers in schuBERT-all for various model sizes. 
	}
	\vspace{-1em}
	\label{fig:heads}
\end{center}
\end{figure}
\begin{figure}[h]
 \begin{center}
	\includegraphics[width=0.9\linewidth]{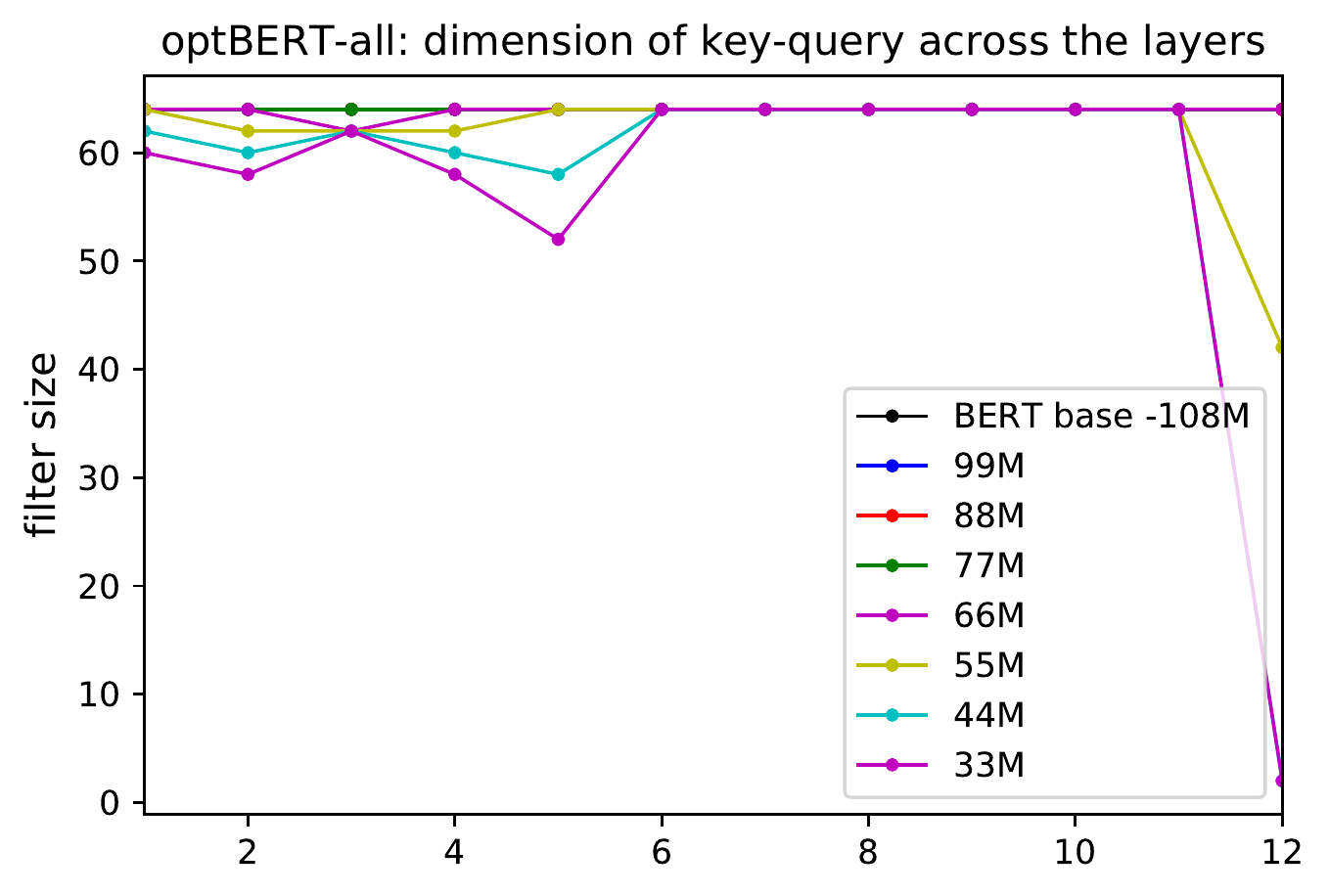}
	\caption{Dimension of key-query vectors across the encoder layers in schuBERT-all for various model sizes.  
	}
	\vspace{-1em}
	\label{fig:key}
\end{center}
\end{figure}
\begin{figure}[h]
 \begin{center}
	\includegraphics[width=0.9\linewidth]{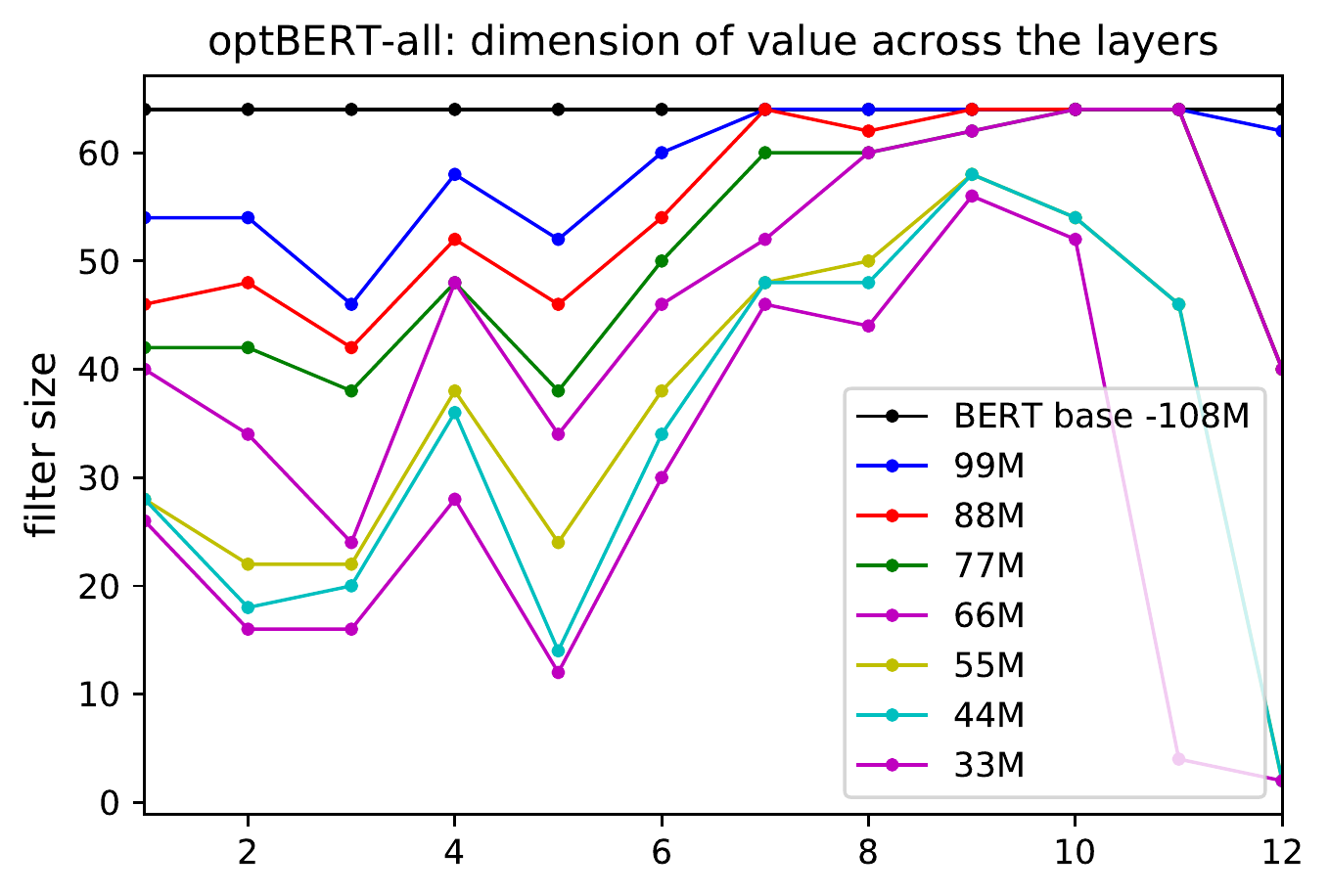}
	\caption{Dimension of value vectors across the encoder layers in schuBERT-all for various model sizes.   
	}
	\vspace{-1em}
	\label{fig:value}
\end{center}
\end{figure}

Figure \ref{fig:heads}, Figure \ref{fig:key} and Figure \ref{fig:value} show the pattern of number of heads, key-query dimension, and value dimension across the encoder layers for various schuBERT-all architectures respectively. There is not much significant pattern in these design dimensions across the layers. The number of attention heads drastically reduce to $1$ in the top layer for very small models. Same is true for key-query and value dimensions. 
Key-query remains almost same as their original value $64$ even when the models are pruned heavily, except in the top layer. Whereas value dimension does decrease significantly from their original value when the models are pruned heavily. 

\end{document}